\documentclass[10pt,twocolumn,letterpaper]{article}

\usepackage{cvpr}              % To produce the CAMERA-READY version
\usepackage{graphicx}
\usepackage{amsmath}
\usepackage{amssymb}
\usepackage{booktabs}
\usepackage{bbding}
\usepackage{caption}
\usepackage{soul}
\usepackage{bbm}
\usepackage{subcaption}
\usepackage{multirow}
\usepackage[accsupp]{axessibility}
\usepackage{xcolor}

\newcommand*{\affaddr}[1]{#1} % No op here. Customize it for different styles.

\newcommand*{\email}[1]{\texttt{#1}}
\usepackage[pagebackref,breaklinks,colorlinks]{hyperref}

\usepackage[capitalize]{cleveref}
\crefname{section}{Sec.}{Secs.}
\Crefname{section}{Section}{Sections}
\Crefname{table}{Table}{Tables}
\crefname{table}{Tab.}{Tabs.}

\def\cvprPaperID{4394} % *** Enter the CVPR Paper ID here
\def\confName{CVPR}
\def\confYear{2024}

\begin{document}

\title{$\textbf{LaRE}^2$: Latent Reconstruction Error Based Method \\ for Diffusion-Generated Image Detection}
% \title{: A generalizable reconstruction-based Reconstruct in the Latent Space: An Efficient and Generalizable Reconstruction-based Framework for Generated-image Detection}

% \and

\author{
Yunpeng Luo \quad Junlong Du \quad Ke Yan\thanks{Corresponding author}
 \quad Shouhong Ding \\
\affaddr{Tencent YouTu Lab} \quad\\
\email{\small \{petterluo, jeffdu, kerwinyan, ericshding\}@tencent.com}
}

\maketitle
%%%%%%%%% ABSTRACT
\begin{abstract}
The evolution of Diffusion Models has dramatically improved image generation quality, making it increasingly difficult to differentiate between real and generated images. This development, while impressive, also raises significant privacy and security concerns. In response to this, we propose a novel \textbf{La}tent \textbf{RE}construction error guided feature \textbf{RE}finement method ($\textbf{LaRE}^2$) for detecting the diffusion-generated images. We come up with the Latent Reconstruction Error (LaRE), the first reconstruction-error based feature in the latent space for generated image detection. LaRE surpasses existing methods in terms of feature extraction efficiency while preserving crucial cues required to differentiate between the real and the fake. To exploit LaRE, we propose an Error-Guided feature REfinement module (EGRE), which can refine the image feature guided by LaRE to enhance the discriminativeness of the feature. Our EGRE utilizes an align-then-refine mechanism, which effectively refines the image feature for generated-image detection from both spatial and channel perspectives. Extensive experiments on the large-scale GenImage benchmark demonstrate the superiority of our $LaRE^2$, which surpasses the best SoTA method by up to \textbf{11.9\%/12.1\%} average ACC/AP across 8 different image generators. LaRE also surpasses existing methods in terms of feature extraction cost, delivering an impressive speed enhancement of \textbf{8 times}. \href{https://github.com/luo3300612/LaRE}{Code is available here}.

% Outperform accuracy?
\end{abstract}

%%%%%%%%% BODY TEXT
\section{Introduction}
\label{sec:intro}
The rapid advancement of Diffusion Models has heralded a new era in the domain of image generation. Through concerted efforts in refining model architecture~\cite{sohl2015deep, dhariwal2021diffusion, rombach2022high}, optimizing training strategies~\cite{nichol2021improved, ho2022classifier}, and enhancing sampling methods~\cite{song2020denoising,lu2022dpm}, contemporary Diffusion Models are now capable of generating images of unprecedented quality, surpassing the boundaries of human imagination. However, this progress raises significant concerns regarding privacy and security associated with the generated images~\cite{juefei2022countering}. The potential for the dissemination of toxic content and misinformation through these images poses a threat to society and could mislead the public. Consequently, there is an urgent need to develop techniques to detect the images generated by these models.

\begin{figure}
    \centering
    \includegraphics[width=0.45\textwidth]{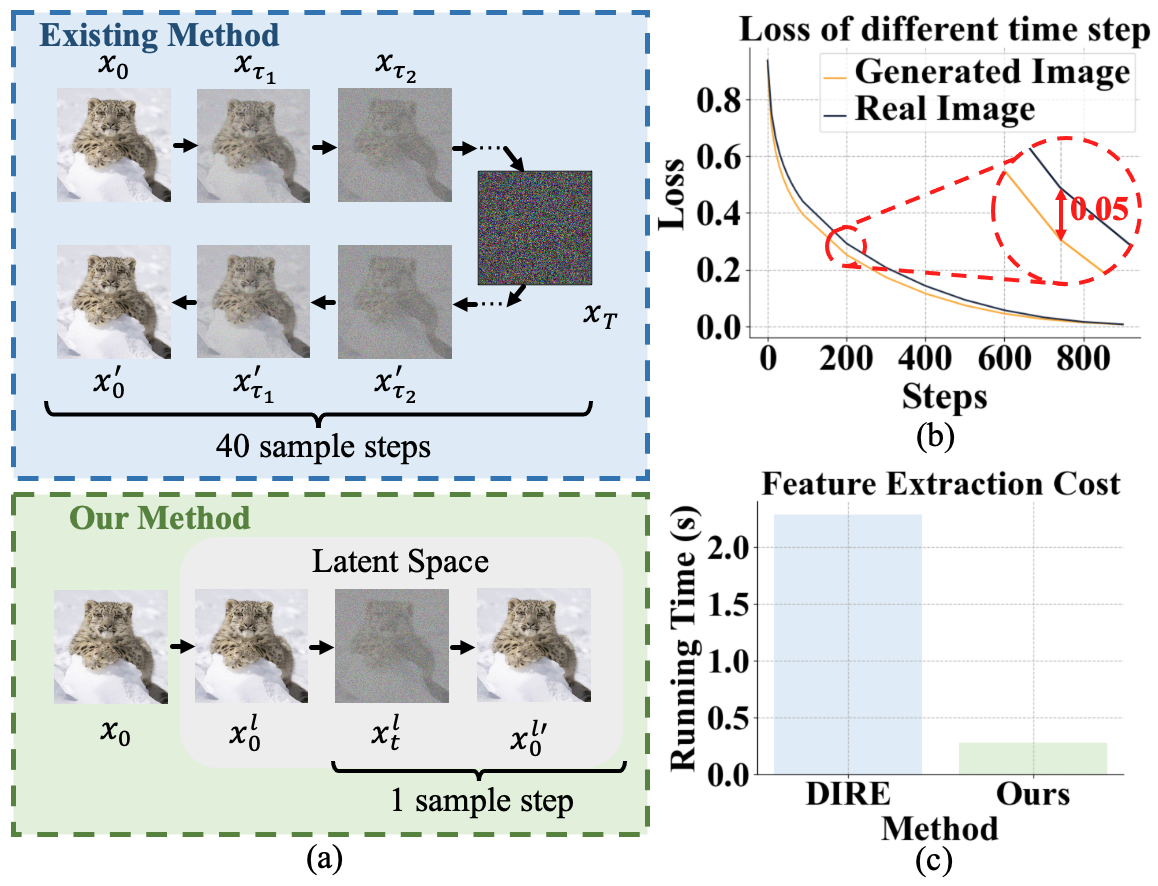}
    \caption{(a) The comparison of reconstruction-based feature extraction. Existing method\cite{wang2023dire} chooses to completely reconstruct an image by first gradually adding noise to the image and then denoising it, which involves dozens of sampling steps. Our method can directly calculate noisy images and denoise them with a single sample step. (b) Statistical analysis of the relationship between the single-step reconstruction loss (1000 images are used) and time step. The obvious gap between the two lines indicates that single-step reconstruction can also reflect the differences between real and generated images. (c) Comparison of the cost of per image feature extraction. Our method is \textbf{8x} faster than DIRE\cite{wang2023dire}. }
    \label{fig1}       % Give a unique label
\end{figure}

A flurry of works\cite{corvi2023detection,ricker2022towards, corvi2023intriguing,wu2023generalizable,wang2023dire,ma2023exposing} has been proposed to detect or study the properties of images generated by diffusion models. Recently, DIRE\cite{wang2023dire} has been proposed to leverage the reconstruction error as a discriminative feature for diffusion-generated image detection. It is based on the assumption that the diffusion-generated images are more easily to be reconstructed by a diffusion model compared with real images. As a result, DIRE shows great cross-model generalizability towards different diffusion models. Based on DIRE, SeDID\cite{ma2023exposing} is also proposed to harness the inherent distributional disparities between naturally occurring and diffusion-synthesized visuals for diffusion-generated image detection.

However, both DIRE and SeDID require multi-step DDIM\cite{song2020denoising} sampling processes in the feature extraction stage, which results in low efficiency for real-world applications. As shown in Fig~\ref{fig1}(c), it takes more than 2 seconds to extract the DIRE feature for an image. Besides, the accumulation loss in the multi-step sampling also introduces uncertainty into the extracted features. Though DDIM\cite{song2020denoising} provides a deterministic inversion method to transform an image into noise, the reliability of this inversion process is not consistently upheld. In addition, DIRE and SeDID use the reconstruction error as the only feature, ignoring the correspondence between the error and the raw image. We therefore ask: (1) Do we need to completely reconstruct the image to get the discriminative feature? (2) Are there better ways to incorporate reconstruction error into image generation detection?

\begin{figure}
    \centering
    \includegraphics[width=0.45\textwidth]{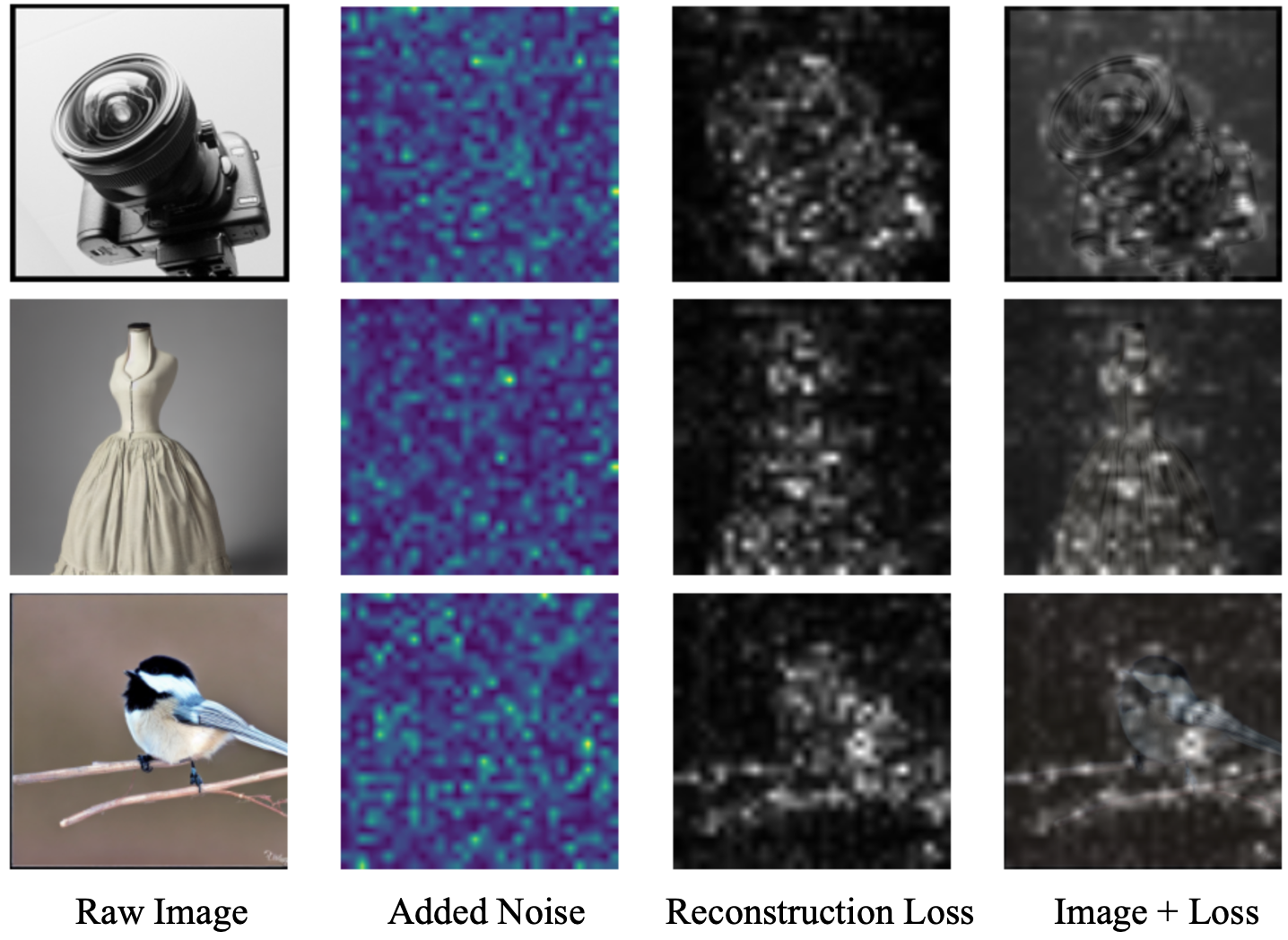}
    \caption{Visualization of reconstruction loss on raw images (i.e. Image + Loss). Though the randomly sampled noises are added to the whole image, there is a trend that the loss in high-frequency regions is typically greater than that in low-frequency regions. }
    \label{fig2}       % Give a unique label
\end{figure}

To address the two questions mentioned above, we conducted a series of exploratory experiments and analyses. For the first question, we investigated the training process of the diffusion model. We found that given any time step $t$, the forward Markov process has a closed-form solution, allowing us to directly transform $x_0$ into $x_t$. Additionally, during the training phase, the model takes $x_t$ and $t$ as inputs and computes the denoising loss, enabling us to obtain the loss through single-step denoising. Therefore, we perform single-step denoising on both 1000 real and 1000 generated images respectively by LDM~\cite{rombach2022high}. The results, as shown in Fig.~\ref{fig1}(b), indicate that even with only single-step reconstruction, the loss of real images is consistently greater than that of generated images. This not only demonstrates that the real images are harder to reconstruct than the generated ones but also suggests that the loss from single-step reconstruction can reflect the differences between real and generated images. To address the second question, we visualize the reconstruction loss on the original image, as shown in Fig.~\ref{fig2}. Upon observation, we have found that the reconstruction loss is positively correlated with the local information frequency of the original image. In the first image, for instance, the reconstruction loss is relatively lower on the low-frequency background, while it is relatively higher on the high-frequency foreground. The same pattern can be observed in other images as well. The reconstruction loss exhibits spatial correlations with the original image, thereby presenting a potential to serve as a valuable cue for the generated image detection.

%

%For spatial refinement, we propose the Error-guided Spatial Refinement module (ESR). It is based on multi-head attention mechanism where we combine the spatially corresponding LaRE and feature map into a pair of keys, thereby conducting Error-guided feature refinement. For channel refinement, we propose the Error-guided Channel Refinement module (ECR) to refine the image through a error-guided gate mechanism
Based on the findings above, we propose a novel Latent REconstruction error guided feature REfinement method ($\text{LaRE}^2$) for diffusion-generated image detection. Our $\text{LaRE}^2$ consists of two parts, the Latent Reconstruction Error (LaRE) and the Error-guided Feature REfinement module (EGRE). LaRE is a more efficient reconstruction feature, which has two improvements compared with existing methods: (1) LaRE is extracted in a single step of the diffusion reverse process, which is significantly more efficient than completely reconstructing the image through dozens of denoising steps. (2) We conduct the reconstruction in the latent space, which further improves the efficiency. Besides, we find that the reconstructed loss is positively correlated with the local information frequency of the original image. Based on that, we come up with the Error-Guided Feature Refinement module (EGRE). In EGRE, we first align LaRE with the image feature map for better correspondings for feature refinement. Then the aligned LaRE is used to refine the image feature in both spatial and channel perspectives. With the guidance of LaRE, EGRE refines the feature map in both spatial and channel perspectives to better reveal the discriminativeness for detecting the generated images.   

To evaluate the effectiveness of our method, we conduct extensive experiments on the GenImage\cite{zhu2023genimage} benchmark, which comprises 2,681,167 images, segregated into 1,331,167 real and 1,350,000 fake images. The fake images are from 8 different generators. As shown in Fig.~\ref{fig1}(c), our LaRE is \textbf{8 times faster} than existing method~\cite{wang2023dire}. In addition, $\text{LaRE}^2$ achieves a significant performance gain by up to \textbf{11.9\%/12.1\%} ACC/AP compared with the best SoTA. The results demonstrated the superiority of our method, which is both effective and generalizable.

The contributions of our work are three-fold:
\begin{itemize}
\item[$\bullet$] \textbf{Novel feature:} We are the first to propose the reconstruction error in latent space for generated-image detection. Compared with the existing method, we remarkably reduce the cost of feature extraction while preserving the essential information required for the detection of diffusion-generated images.
\item[$\bullet$] \textbf{Novel module:} We qualitatively analyze the reason for the effectiveness of the reconstruction loss. Based on that, we come up with a novel module EGRE, which conducts an Error-guided feature refinement to enhance the discriminativeness of image features.
\item[$\bullet$] \textbf{Superior performance:} Extensive experiments demonstrate the effectiveness of our method. We achieve \textbf{11.9\%/12.1\%} ACC/AP gain on the large-scale GenImage benchmark, significantly outperforming the SoTA methods.
\end{itemize}

\section{Related Works}
\label{sec:related}
% 写的啥玩意
% 注意表述一致性：deep generated image detection
% 表述高端一点，废话少一点
\subsection{Diffusion Model}
Diffusion Models\cite{ho2020denoising, sohl2015deep, yang2022diffusion} have become new state-of-the-art deep generative models. Inspired by nonequilibrium thermodynamics, Denoising Diffusion Probabilistic Models (DDPM) demonstrate promising generative quality, which inspires a flurry of studies on diffusion models\cite{song2020denoising, dhariwal2021diffusion, rombach2022high, baranchuk2021label, tang2023emergent, zhao2023unleashing, xu2023open, clark2023text}. DDIM\cite{song2020denoising} generalize DDPMs~\cite{ho2020denoising} via a class of non-Markovian diffusion processes to accelerate sampling. ADM~\cite{dhariwal2021diffusion} firstly obtain better generation quality than GANs~\cite{goodfellow2020generative} on ImageNet~\cite{deng2009imagenet}. Latent Diffusion Models~\cite{rombach2022high} apply the diffusion process in the latent space to improve the efficiency of diffusion models, and enable diffusion models with text conditioning inputs through the cross-attention conditioning mechanism. There are also works that employ pre-trained diffusion models for various downstream tasks~\cite{baranchuk2021label, tang2023emergent, zhao2023unleashing, xu2023open, clark2023text}. The rapid advancement of deep generative models has given rise to concerns regarding the potential for malicious utilization of the generated images~\cite{juefei2022countering}. Consequently, it is urgent to develop robust techniques to detect generated images. 

\subsection{Deep Generated Image Detection}
% deepfake 针对GAN的方法介绍，说明这些方法现在都不足以应对了
Deep-generated image detection has achieved significant improvements due to the contributions of previous research. Initially, researchers attempted to use hand-crafted features including color cues~\cite{mccloskey2018detecting}, saturation cues~\cite{mccloskey2019detecting}, blending~\cite{li2020face} artifacts, co-occurrence features~\cite{nataraj2019detecting}. Then CNN is leveraged to detect the generated images\cite{liu2020global, wang2020cnn, marra2018detection}. Some works\cite{zhang2019detecting, frank2020leveraging} also find that there are obvious visual artifacts in the GAN-generated images and detect these images from the frequency view~\cite{qian2020thinking}. Most of the above works are developed and designed for GAN-generated images. With the advance of diffusion models, many methods are also provided for the study of diffusion-generated image detection. Corvi \emph{et. al}\cite{corvi2023detection,ricker2022towards} find that state-of-the-art detectors developed for GAN suffer from a severe performance drop when applied to Diffusion-generated images. Corvi \emph{et. al}\cite{corvi2023intriguing, corvi2023detection, ricker2022towards} also find that the spectrum artifacts can also rise up in the Diffusion-generated images. Wu \emph{et. al}\cite{wu2023generalizable} formulate the synthetic image detection as an identification problem and achieve generalizable detection through language-guided contrastive learning. DIRE\cite{wang2023dire} and SeDIE\cite{ma2023exposing} are proposed to leverage reconstruction error from the diffusion model to achieve generalizable diffusion-generated image detection. Different from these methods, we are the first to show that reconstruction error in the latent space can also benefit diffusion-generated image detection, which is more efficient and generalizable.

\begin{figure}
    \centering
    \includegraphics[width=0.47\textwidth]{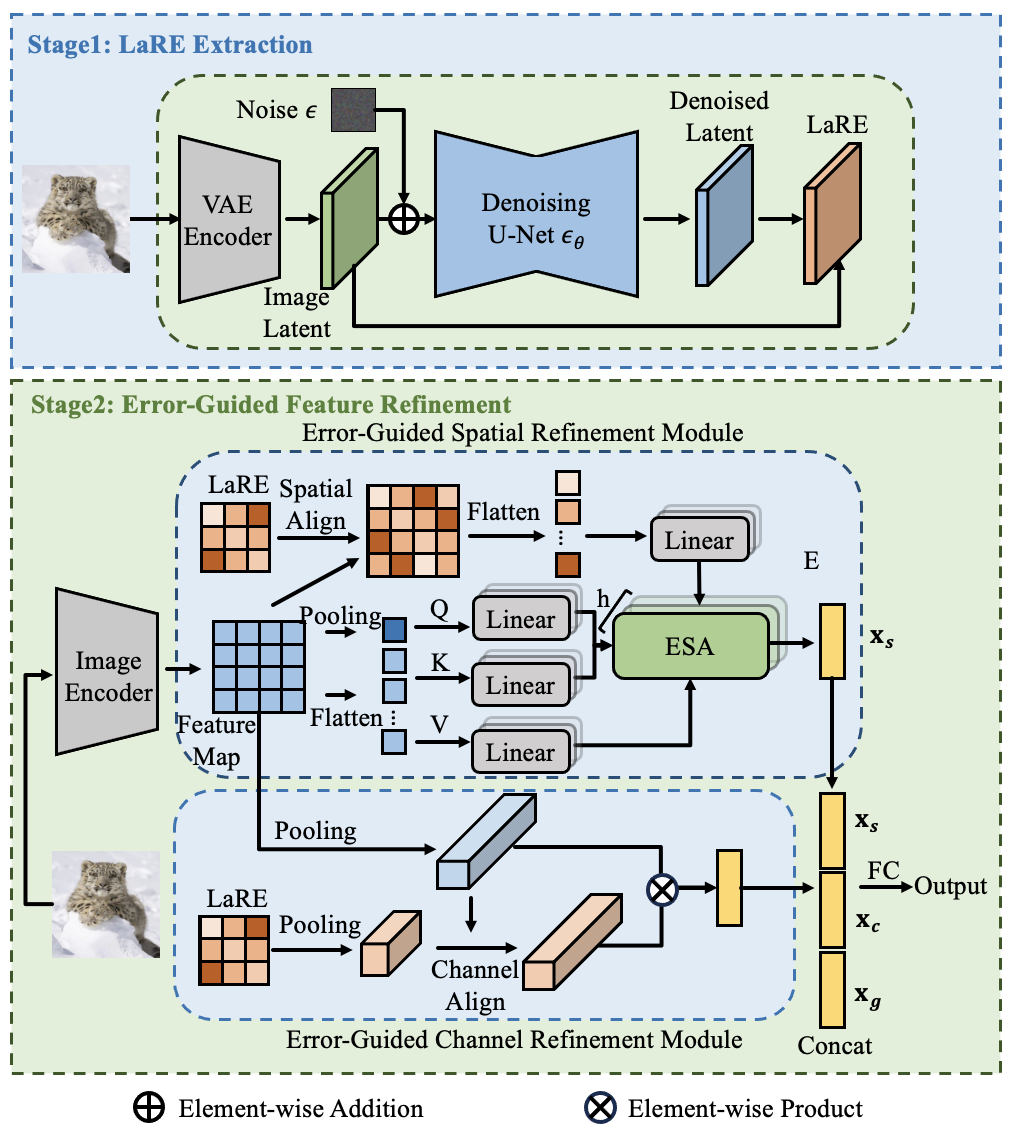}
    \caption{Overview of our method. In the first stage, we extract LaRE in the latent space through single-step reconstruction. In the second stage, to exploit LaRE, we propose the Error-guided Feature Refinement Module, which consists of the Error-guided spatial refinement module and the Error-guided Channel Refinement module. From both spatial and channel perspectives, LaRE is used to enhance the discriminativeness of the image feature for generated image detection. }
    \label{fig:overall}       % Give a unique label
\end{figure}

% \begin{equation}
%     L=\frac{1}{N}\sum_{i=1}^{N}-\log\frac{\sum_{j=1,j\neq i}^N\mathbbm{1}(y_i=y_j)\exp{(\mathbf{I}_i\cdot \mathbf{I}_j})}{\sum_{j=1,j\neq i}^N\exp{(\mathbf{I}_i\cdot \mathbf{I}_j})}
% \end{equation}
\section{Preliminaries}
Diffusion Models has achieved remarkable image generation performance. Typically, it involves two Markov processes. In the forward process, Gaussian noise is gradually added to the raw image $\mathbf{x}_0$ until the image is asymptotically transformed to pure noise, which is defined as:

\begin{equation}
    q(\mathbf{x}_t|\mathbf{x}_{t-1})=N(\mathbf{x}_t;\sqrt{\frac{\alpha_t}{\alpha_{t-1}}}\mathbf{x}_{t-1},(1-\frac{\alpha_t}{\alpha_{t-1}}\mathbf{I})),
\end{equation}
where $x_t$ is the noisy image at step $t$ and $\alpha_t$ is predefined noise schedule. According to the properties of Markov process and Gaussian distribution, we can get $\mathbf{x}_t$ from $\mathbf{x}_0$ directly by:
\begin{equation}
        q(\mathbf{x}_t|\mathbf{x}_{0})=N(\mathbf{x}_t;\sqrt{\alpha_t}\mathbf{x}_{0},(1-\alpha_t)\mathbf{I})).
        \label{Eq:forward_close_form}
\end{equation}
In the reverse process, the noisy image is gradually de-noised to get the raw image, which is defined as:
\begin{equation}
    p_\theta(\mathbf{x}_{t-1}|\mathbf{x}_{t})=N(\mathbf{x}_{t-1};\mathbf{\mu}_\theta(\mathbf{x}_t,t),\mathbf{\Sigma}_\theta(\mathbf{x}_t,t)),
\end{equation}
where $p_\theta$ is typically parameterized by neural networks. During training, a neural network $\epsilon_\theta$ is trained to predict the added noise $\epsilon$, given the noisy image $x_t$ and corresponding time step $t$:
\begin{equation}
    L_\theta(\mathbf{x}_0, t) = \Vert\epsilon-\epsilon_\theta(\sqrt{\bar{\alpha}}_t\mathbf{x}_0+\sqrt{1-\bar{\alpha}_t}\epsilon,t)\Vert^2.
\label{Eq:loss}
\end{equation}
During training, $t$ and $\epsilon$ are randomly sampled.

\section{Methods}
In this section, we elaborate on each component of our $\text{LaRE}^2$, as illustrated in Fig.~\ref{fig:overall}.
% Specifically, we first present the Latent Reconstruction Error (LaRE) in Sec.~\ref{LaRE}. Then, we introduce our Error-Guide Feature Refinement (EGRE) module in Sec.~\ref{EGRE}. 

\subsection{Latent Reconstruction Error}
\label{LaRE}
Recent methods\cite{wang2023dire} propose to use reconstruction error as the feature for generated-image detection. Through DDIM~\cite{song2020denoising} inversion, an image is first inversed to noise and then re-generated based on the noise. However, this method has several limitations: (1) It is relatively slow to inverse and re-generate an image. For example, it takes more than 2 seconds to extract DIRE for an image on a Tesla V100 GPU as shown in Fig.~\ref{fig1}(c), which limits its realistic applications; (2) The reconstruction of the image is not always reliable. Error could accumulate during the Markov forward and reverse process, making it hard to tell whether the bad reconstruction is from the difference between real and fake images, or the reconstruction itself. 

% 这边需要强调好
% TODO 探索性实验的证据需要加上
Here we come up with a new reconstruction feature, named Latent Reconstruction Error (LaRE). LaRE is based on the assumption that if the generated images can be completely reconstructed with relative ease~\cite{wang2023dire}, then they can also be more easily reconstructed at every single step of the reverse diffusion process.

For an image $x$, we first get its latent code $x_0$ by VAE. Then we calculate the LaRE by:

\begin{equation}
    \begin{aligned}
    \mathbf{L}_\epsilon = \epsilon-\epsilon_\theta(\sqrt{\bar{\alpha}}_t\mathbf{x}_0+\sqrt{1-\bar{\alpha}_t}\epsilon,t)
    \end{aligned}
\end{equation}

\begin{equation}
    \text{LaRE}= \mathbbm{E}_\epsilon\left[\mathbf{L}_\epsilon\odot\mathbf{L}_\epsilon\right].
    \label{Eq:Lare1}
\end{equation}

As shown in Fig.~\ref{fig:overall}, Our LaRE takes advantage of two special properties of Diffusion Models. (1) The forward process has a close form. According to Eq.~\ref{Eq:forward_close_form}, we can get $\mathbf{x}_t$ directly from $\mathbf{x}_0$; (2) Diffusion Model is trained to denoise a noisy image from any time step $t$, as shown in Eq.~\ref{Eq:loss}. Therefore, given $\mathbf{x}_t$, $t$, and $\epsilon$, we can directly get the reconstruction error by only single-step denoising, which is much more efficient than completely reconstructing an image. Compared with complete image reconstruction, our LaRE has several advantages: (1) Only single-step denoising is required to get LaRE, which significantly improves its efficiency. (2) Reconstruction is conducted in the latent space, which is also more efficient. (3) No need for complete inversion and reconstruction of the image, which eliminates the potential accumulation of errors that may occur during multiple inversion and reconstruction steps. 

With efficiency, the easier reconstruction assumption still holds for our LaRE. Recall that a diffusion-generated image is more easily reconstructed by diffusion models\cite{wang2023dire}. So the reconstruction error should be relatively small when completely reconstructing an image through the whole reverse process. That implies the reconstruction error is also relatively small during each step of the reverse process, which is also validated in Fig.~\ref{fig1}(b).
We leverage the Monte Carlo method to estimate LaRE in Eq.~\ref{Eq:Lare1}: 
\begin{equation}
    \text{LaRE} = \frac{1}{e}\sum_i^e\left(\mathbf{L}_{\epsilon_i}\odot \mathbf{L}_{\epsilon_i}\right),
    \label{Eq:Lare2}
\end{equation}
where $e$, $t$ are pre-defined hyperparameters. Specifically, we conduct denoising for $e$ times and average the denoising loss to extract LaRE.
\subsection{Error-Guided Feature Refinement}
\label{EGRE}
% 探索性实验的证据需要加上
Existing methods exploit the reconstruction error as the only feature to detect the generated image, which ignores the relationships between the reconstruction error and the raw image. To better understand how reconstruction error helps generated-image detection, we visualize the extracted LaRE on the raw image, as shown in Fig.~\ref{fig2}. We find several interesting conclusions. 1) The loss norm is proportional to the noise norm. It is easy to understand that the stronger the added noise is, the harder to reconstruct; 2) The loss norm is also proportional to the frequency of local image patches. That the low-frequency patches are easier to reconstruct than the high-frequency ones. Therefore, the reconstruction performance on the high-frequency parts of the image plays an important role in the generated image detection. To apply this valuable information, we come up with a novel Error-guided feature REfinement Module (EGRE). Based on the reconstruction error, our EGRE can better refine the image feature to reveal the discriminative parts for generated image detection.

\begin{figure*}
    \centering
    \includegraphics[width=1\textwidth]{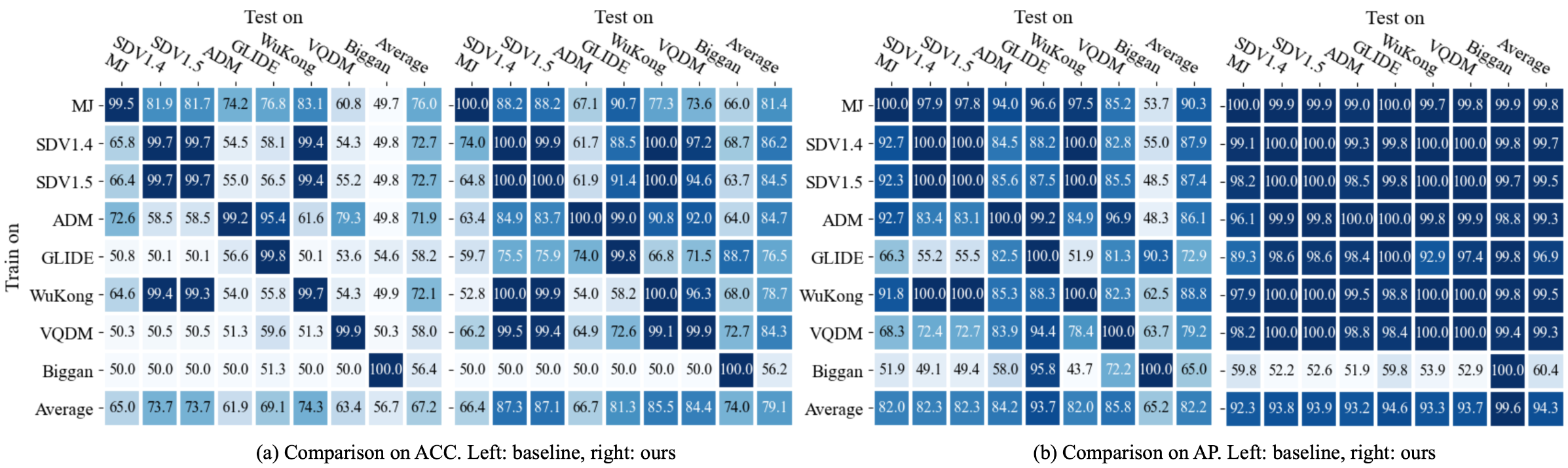}
    \caption{Results of cross-validation on different training and testing subsets. For each generator, we train a model and test it on all 8 generators. For both DIRE\cite{wang2023dire} and our method. accuracy (ACC) and average precision(AP) are reported. }
    \label{fig:alleval}       % Give a unique label
\end{figure*}

\subsubsection{Error-guided Spatial Refinement Module}
We propose the Error-guided Spatial Refinement module (ESR) for feature refinement from the spatial perspective. First, we spatially align LaRE and the image feature map by a simple adaptive average pooling layer, so that they have the same spatial size. Now we have the spatially aligned input feature map $\mathbf{x}\in\mathbbm{R}^{HW\times C_1}$ and LaRE $\mathbf{e}\in\mathbbm{R}^{HW\times C_2}$. $\bar{\mathbf{x}}\in\mathbbm{R}^{1\times C_1}$ is the global feature, which is the average of $\mathbf{x}$. To refine the feature from the spatial perspective, we choose an Error-guided Spatial Attention module (ESA): 
\begin{equation}    
    \text{ESA}(\mathbf{Q},\mathbf{K},\mathbf{V},\mathbf{E}) = \text{softmax}(\frac{\mathbf{Q}\mathbf{K}^T}{\sqrt{d_k}}+\mathbf{E})\mathbf{V}.
\end{equation}
To further enhance the feature representation, we adopt a multi-head mechanism from Transformer\cite{vaswani2017attention}. Therefore, our Multi-Head Error-guided Spatial Attention module (MHESA) is defined as:
\begin{equation}
    \begin{aligned}
    \text{MHESA}(\mathbf{Q},\mathbf{K},\mathbf{V},\mathbf{E}) =\text{Concat}(\text{head}_1,...,\text{head}_h)\mathbf{W}^O, \\
    \text{where}\,\, \text{head}_i =\text{ESA}(\mathbf{Q}\mathbf{W}_i^Q, \mathbf{K}\mathbf{W}_i^{K}, \mathbf{V}\mathbf{W}_i^{V}, (\mathbf{E}\mathbf{W}_i^E)^T).
    \end{aligned}
\end{equation}
where $h$ is the number of heads, $\mathbf{W}_i^Q\in \mathbbm{R}^{C_1\times d}$, $\mathbf{W}_i^K\in \mathbbm{R}^{C_1\times d}$, $\mathbf{W}_i^V\in \mathbbm{R}^{C_1\times d}$,$\mathbf{W}_i^E\in \mathbbm{R}^{C_2\times 1}$ are learnable projection matrices, $d$ is the hyperparameter. The spatial refined feature $x'$ is calculated by:
\begin{equation}
    \mathbf{x}_s = \text{MHESA}(\bar{\mathbf{x}}, \mathbf{x}, \mathbf{x}, \mathbf{e}),
\end{equation}
where $\bar{\boldsymbol{x}}$ is the only query attending to the feature map $\boldsymbol{x}$. In this stage, the aligned LaRE is used to re-weight the attention score of scaled dot-product attention. Since LaRE and the feature map are spatially aligned, LaRE can emphasize important information from the spatial perspective. 

\subsubsection{Error-guided Channel Refinement Module}
For channel refinement, we first squeeze both $\mathbf{x}$ and $\mathbf{e}$ to $\bar{\mathbf{x}}\in\mathbbm{R}^{1\times C_1}$ and $\bar{\mathbf{e}}\in\mathbbm{R}^{1\times C_2}$. Then our channel refinement is achieved by a gate mechanism:
\begin{equation}
    \mathbf{x}_c = \text{sigmoid}\left(\bar{\mathbf{e}}\mathbf{W}\right)\odot\bar{\mathbf{x}},
\end{equation}
where $\mathbf{W}\in\mathbbm{R}^{C1\times C2}$ are learnable parameters to align LaRE and the feature map from the channel perspective. Finally, $\mathbf{x}_s$, $\mathbf{x}_c$ and the original global feature $\mathbf{x}_g$ from the last convolution block are concated as the final feature $\mathbf{x}_{\text{EGRE}}$.
\begin{equation}
    {\mathbf{x}_{\text{EGRE}} = \text{Concat}(\mathbf{x}_s, \mathbf{x}_c, \mathbf{x}_g)}.
\end{equation}
$\mathbf{x}_{\text{EGRE}}$ is followed by one FC layer. We train our model with the binary cross entropy loss.

\section{Experiments}
\subsection{Datasets and Evaluation Metrics}
We evaluate our proposed method using the GenImage~\cite{zhu2023genimage} dataset. This dataset comprises a total of 2,681,167 images, divided into 1,331,167 real images and 1,350,000 generated images. The generated images are from 8 different generative models, namely BigGAN~\cite{brock2018largebiggan}, GLIDE~\cite{nichol2021glide}, VQDM~\cite{gu2022vectorvqdm}, Stable Diffusion V1.4\&V1.5~\cite{rombach2022high}, ADM~\cite{dhariwal2021diffusion}, Midjourney~\cite{Midjourney}, and Wukong~\cite{wukong}. All of the fake images are generated using the template prompt ``photo of CLS", where ``CLS'' is replaced by one of the 1000 labels from ImageNet\cite{deng2009imagenet}. These images are split into 8 subsets where each subset contains partial real images and all the fake images generated by one of the above generators. We adhere to the official division of the dataset in our research, allocating 2,581,167 images for training and reserving the remaining 100,000 images for validation. Following DIRE~\cite{wang2023dire}, we employ Accuracy~(ACC) and Average Precision~(AP) as our evaluation metrics. More details and results are included in our supplementary materials.

\begin{table*}[t]
	\centering
\begin{tabular}{l|cccccccc|c}
\toprule[1pt]
\multirow{2}{*}{Methods} & \multicolumn{8}{c|}{Testing Subset}                                                                                           & \multirow{2}{*}{\begin{tabular}[c]{@{}c@{}}Avg\\ ACC.(\%)\end{tabular}} \\ \cline{2-9}
                        & Midjourney    & SDV1.4        & SDV1.5        & ADM           & GLIDE         & Wukong        & VQDM          & BigGAN        &                                                                         \\ \hline
CNNSpot\cite{wang2020cnnspot}                 & 58.2          & 70.3          & 70.2          & 57.0          & 57.1          & 67.7          & 56.7          & 56.6          & 61.7                                                                    \\
Spec\cite{zhang2019detectingSPEC}                    & 56.7          & 72.4          & 72.3          & 57.9          & 65.4          & 70.3          & 61.7          & 64.3          & 65.1                                                                    \\
F3Net\cite{qian2020thinkingF3Net}                   & 55.1          & 73.1          & 73.1          & 66.5          & 57.8          & 72.3          & 62.1          & 56.5          & 64.6                                                                    \\
GramNet\cite{liu2020globalGramNet}                 & 58.1 & 72.8          & 72.7          & 58.7          & 65.3          & 71.3          & 57.8          & 61.2          & 64.7                                                                    \\
DIRE\cite{wang2023dire}\dag                    & 65.0          & 73.7          & 73.7          & 61.9          & 69.1          & 74.3          & 63.4          & 56.7          & 67.2                                                                      \\ \hline
Ours           & \textbf{66.4} & \textbf{87.3} & \textbf{87.1} & \textbf{66.7} & \textbf{81.3} & \textbf{85.5} & \textbf{84.4} & \textbf{74.0} & \textbf{79.1}                                                           \\ \toprule[1pt]
\end{tabular}
    \caption{Performance comparisons on GenImage test set. For each data point, eight models are trained on eight generators respectively. Then eight models are tested on the specified test sets and the accuracy scores are averaged. \dag indicates our reproduction. }\smallskip
    \label{table1}
\end{table*}

\subsection{Implementation Details}
All experiments are conducted using the GenImage dataset. To obtain the LaRE, we employed Stable Diffusion V1.5\cite{rombach2022high} with a step size of $t=200$ and a noise ensemble size of $e=4$. Images are resized to $256\times256$ for LaRE extraction. We use the prompt ``a photo'' for all images. For both training and testing phases, images are resized to $224\times224$. Our model utilizes the CLIP\cite{radford2021learningclip} pretrained ResNet50\cite{he2016deepresnet} as the backbone architecture. We set the batch size to 48 and the learning rate to $1e^{-4}$. Training is performed on a single Nvidia Tesla V100 GPU. We train eight models on eight subsets respectively, each corresponding to a different generation method. Model selection is based on performance in the validation set, which shares the same generation method as the training set. 

\subsection{Cross-Generator Image Classification}
We evaluate the performance of our model trained by one of the subsets and test on all eight subsets. The results are shown in Fig.~\ref{fig:alleval}. Several conclusions can be made from the results. (1) It is easy to detect generated images from seen generators. Both the DIRE and our proposed method demonstrate superior performance when training and testing images share the same generator, as shown by the diagonal of each image. (2) It is challenging to generalize the model to unseen generators. However, it is relatively easier to generalize the model to similar generator structures than to different structures. For example, a model trained on the BigGAN subset achieves bad performance when tested on other subsets, which are all diffusion-based generators. While the model trained on the SDV1.5 subset achieves relatively better performance on other diffusion-based models. (3) Our model is more robust and generalizable than DIRE. Our model trained on diffusion-based generators achieves better generalizability on both gan-based and diffusion-based generators. This indicates that our $\text{LaRE}^2$ can better capture discriminative features to detect the generated images. 

\subsection{Compare with the State of the Arts}
In this section, we compare our method with several state-of-the-art generated image detection methods, including CNNSpot\cite{wang2020cnnspot}, Spec\cite{zhang2019detectingSPEC}, F3Net\cite{qian2020thinkingF3Net}, GramNet\cite{liu2020globalGramNet} and DIRE\cite{wang2023dire}. In accordance with the GenImage benchmark, eight models are trained on each subset. Subsequently, these models are evaluated on the specified test set, and their accuracy scores are averaged to produce a data point for inclusion in the table. This benchmark presents a formidable challenge, as the overall accuracy is closely tied to the cross-generator generalizability of the proposed techniques. As illustrated in Tab.~\ref{table1} and Fig.~\ref{fig:alleval}, our approach attains state-of-the-art performance across all eight generators. Notably, our overall accuracy exhibits a substantial improvement, surpassing the previous state-of-the-art by 11.9\%/12.1\% ACC/AP. This result demonstrates the superior generalizability of our method.

\subsection{Ablation Study}
In this section, we conduct several ablative studies to quantify the contribution of each design in our model and the influence of the hyperparameters. 

\begin{table*}[t]
	 \centering
% \resizebox{0.48\textwidth}{!}{
\small
\renewcommand\arraystretch{1.2}
\setlength{\tabcolsep}{1mm}{

\begin{tabular}{ccccccccccccc}
\toprule[1pt]
Model & ESR                       & ECR                       & CLS                & MJ        & SDV1.4      & SDV1.5      & ADM       & GLIDE     & WuKong      & VQDM       & Biggan    & Avg       \\ \hline
A     &                           &                           &                           & 52.9/94.4 & 100.0/100.0 & 100.0/100.0 & 55.0/89.5 & 75.5/98.7 & 99.7/100.0  & 51.8/84.4  & 50.4/71.7 & 73.1/92.3 \\
B     & \checkmark &                           &                           & 61.8/96.9 & 100.0/100.0 & 100.0/100.0 & 60.3/95.4 & 87.2/99.1 & 100.0/100.0 & 86.2/94.6  & 58.1/95.0 & 81.7/97.6 \\
C     &                           & \checkmark &                           & 57.4/95.3 & 100.0/100.0 & 100.0/100.0 & 57.2/91.6 & 81.2/98.9 & 100.0/100.0 & 78.3/89.8  & 56.2/82.3 & 78.8/94.7 \\
D     & \checkmark & \checkmark &                           & 64.8/98.2 & 100.0/100.0 & 100.0/100.0 & 61.9/98.5 & 91.4/99.8 & 100.0/100.0 & 94.6/100.0 & 63.7/99.7 & 84.5/99.5 \\
E     & \checkmark & \checkmark & \checkmark & 66.2/99.3 & 100.0/100.0 & 100.0/100.0 & 64.5/99.5 & 91.5/99.9 & 100.0/100.0 & 97.7/100.0 & 67.4/99.8 & 85.9/99.8 \\
\toprule[1pt]
\end{tabular}

}
% }
    \caption{Ablative studies on using different modules or prompts. `CLS' indicates to use class-specific prompt for LaRE extraction.}
    \label{table:ablation}
\end{table*}

\subsubsection{Influence of EGRE}
 The contribution of individual components in EGRE to the overall performance of the method is reported in Tab.~\ref{table:ablation}. Starting with the most simple baseline, \emph{i.e.}, training without LaRE (Model A in the table), each component is added building up to the proposed approach. We measure the effect of the Error-guided Spatial Refinement module (ESR) and Error-guided Channel Refinement module (ECR). Specifically, ESR achieves 8.6\%/5.3\% ACC/AP gain and ECR achieves 5.7\%/2.4\% ACC/AP gain compared with the baseline. Both ESR and ECR significantly improve the performance, demonstrating the effectiveness of the two modules. Relatively, ESR brings a higher improvement in metrics compared with ECR, indicating a greater benefit from spatial feature refinement. This is also attributed to the better alignment of LaRE with the original image at the spatial scale as shown in Fig.~\ref{fig2}. By combining ESR and ECR, our approach achieves a further improvement of 11.4\%/7.2\% ACC/AP. By leveraging LaRE for feature refinement at both spatial and channel scales, our model achieves more robust and generalizable image generation detection performance.

\begin{figure}
    \centering
    \includegraphics[width=0.45\textwidth]{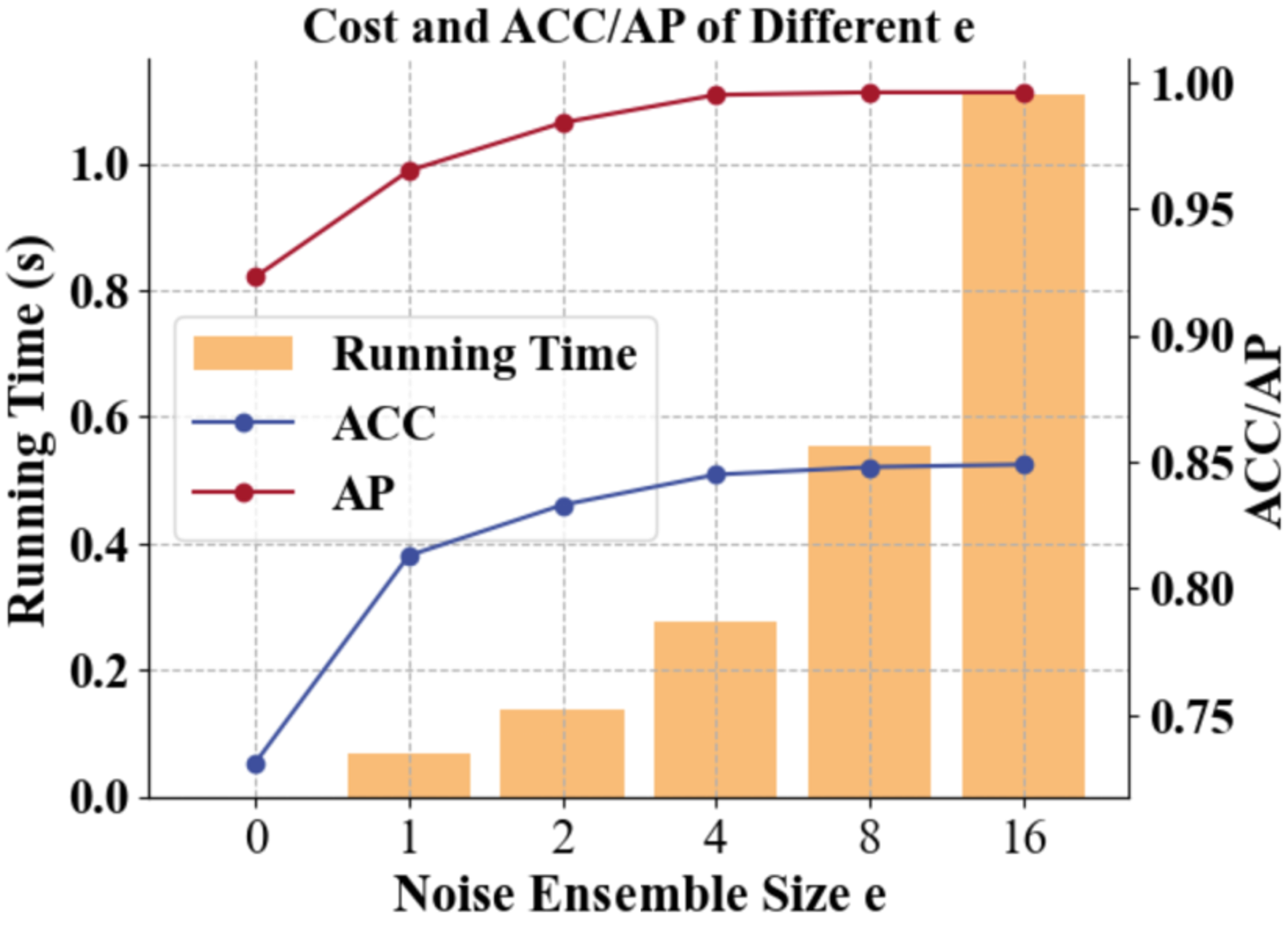}
    \caption{Trade-off between detection performance and feature extraction cost. When e=4, the model achieves the best trade-off.}
    \label{fige}       % Give a unique label
\end{figure}

\subsubsection{Influence of Noise Ensemble}
We systematically evaluated the impact of the noise ensemble parameter $e$ on both the accuracy and feature extraction runtime. As shown in Fig.~\ref{fige}. The results demonstrate a consistent increase in accuracy as $e$ increased. However, this improvement in accuracy comes at the cost of increased feature extraction runtime. Upon careful consideration of the trade-off between accuracy and computational cost, we propose $e=4$ as the optimal choice. This selection of $e$ provides an effective balance, maintaining the performance of our model while ensuring computational efficiency.

\subsubsection{Influence of Prompts}
Because Latent diffusion is a text-conditioned model, a prompt is needed to denoise an image. Here we try to use a class-specific prompt instead of an unchanged prompt. We choose the template ``a photo of CLS'' where ``CLS'' is replaced with the ImageNet label of the image. Results are shown in Tab.~\ref{table:ablation}. When using class-specific prompts (Model E), we can achieve further improvements by 1.4\%/0.3\% ACC/AP. This suggests that the category information of images can aid in the extraction of LaRE features. Even in the absence of category information, it is interesting that the accuracy does not decline significantly. We speculate this is because we did not fully convert the images into noise. Even with the added noise, the image still retains some of the original important semantic information, which can assist the model in denoising and thus obtaining good LaRE features. This experiment validates that when the textual information of an image is known, the extracted LaRE features can further enhance the performance. However, we still use the generic prompt in our method, as it already achieves satisfactory results. As for accurately describing images to further enhance performance, we leave that for future research work.

\begin{figure}
    \centering
    \includegraphics[width=0.40\textwidth]{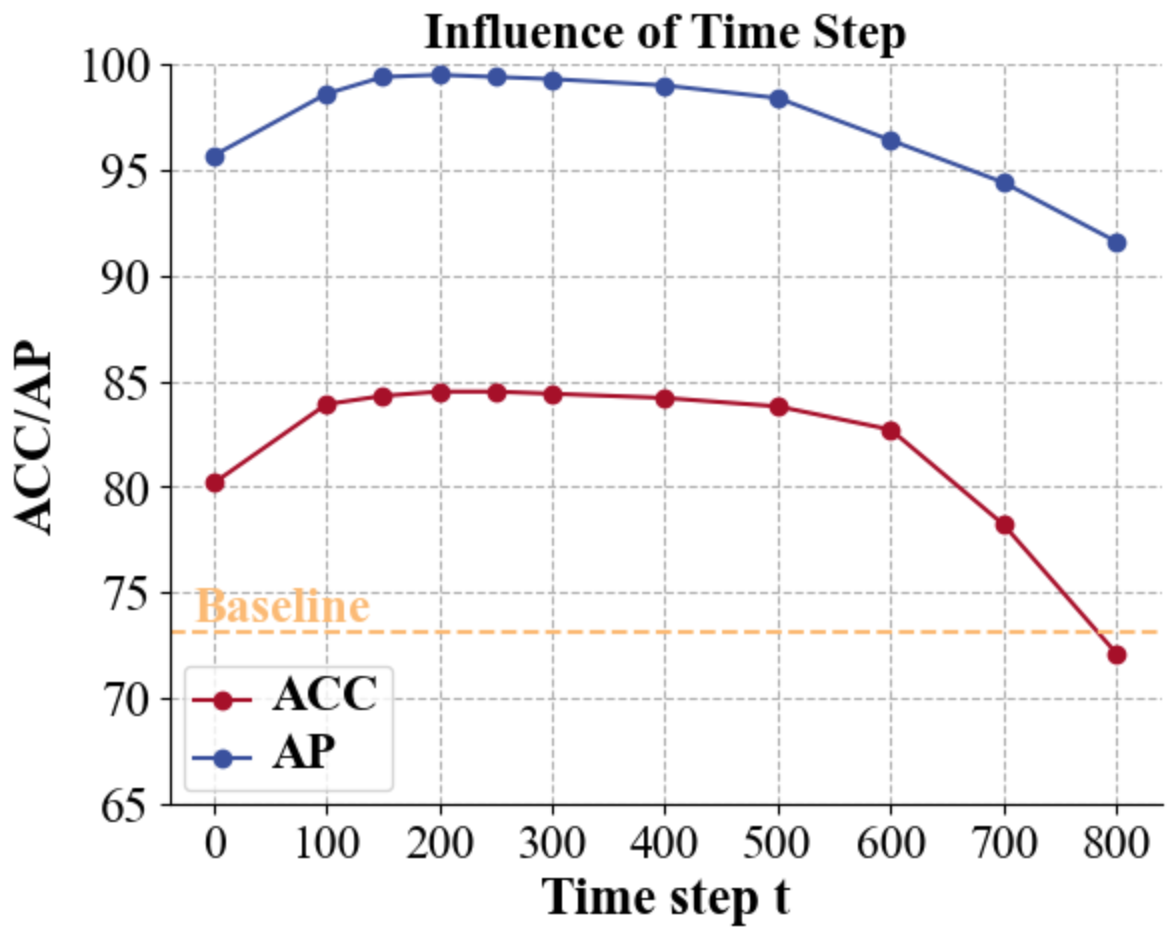}
    \caption{ACC and AP when choosing different $t$. The results demonstrate that our model is robust to the choice of $t$. }
    \label{fig:timestep}       % Give a unique label
\end{figure}

\begin{figure}
    \centering
    \includegraphics[width=0.45\textwidth]{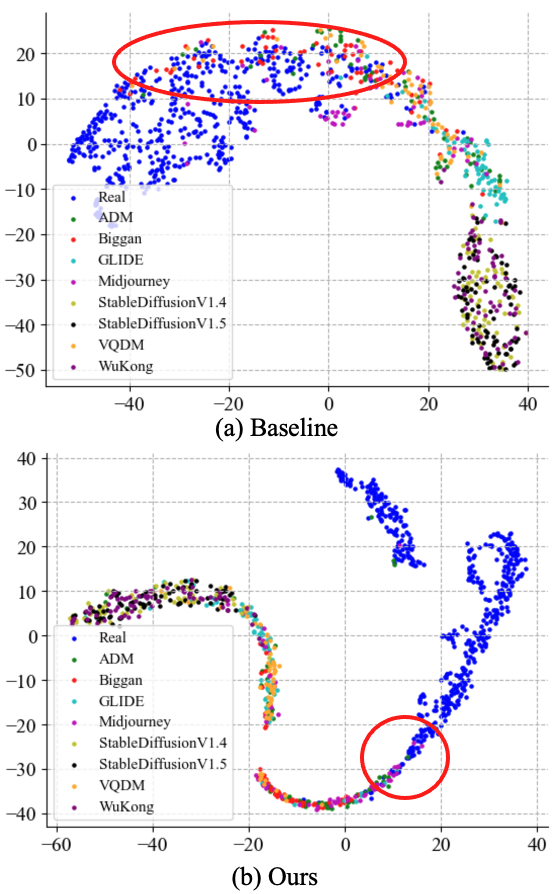}
    \caption{Visualization of the spatial representation of the baseline model (a) and our model (b). Both models were trained on the SDV1.5 subset. In the case of seen data, both models exhibit good discrimination in the feature space. However, for unseen data, the baseline model shows a significant overlap between real images and fake images (red circle), whereas our model exhibits a smaller overlap (red circle). This further indicates the better generalization capability of our model.}
    \label{figfs}       % Give a unique label
\end{figure}

\subsubsection{Influence of Sample Step}
Here we conduct the ablative experiment on the sample step $t$. As illustrated in Fig.~\ref{fig1}(b), there is a gap between the reconstruction loss of real images and the generated images. When $t\in [100, 500]$, there is a more obvious difference in the reconstruction loss between the two types of images. To verify the impact of $t$ selection on the performance, we conducted experiments with different values of $t$, and the results are shown in Fig.~\ref{fig:timestep}. When $t\in [150,300]$, the model exhibits stable performance, indicating the robustness of our model with respect to $t$. However, when t exceeds this range, the performance of the model tends to decline.

\subsubsection{Influence of Input Information}
We conduct experiments to train the detection model with different inputs. The results are shown in Tab.~\ref{table3}. Since our LaRE is a compressed latent representation whose size is only $32\times 32\times4$. We train a Resnet-20 with LaRE as the only input (Model B). The experimental results show that both the AP and ACC metrics have decreased significantly. We speculate that there are two reasons. Firstly, LaRE is in the latent space and has more information loss compared to the original image. It is not sufficient to serve as the only information for detecting the generated images. Secondly, as shown in the Fig.~\ref{fig2}, LaRE mainly reflects the reconstruction quality of the original image at different spatial positions. It is better to leverage LaRE to enhance the representation of the original image features. Therefore, we conducted further experiments. In model C, we only concatenated LaRE with the feature map of the image, which also improves the performance. This validates our hypothesis. Besides, model D achieves significant performance gain compared with Model C. By using EGRE, we can further improve the performance of our model by a significant margin, which also confirms the advantages of EGRE over simple feature concatenation.

\begin{table}[t]
	\centering
 \setlength{\tabcolsep}{1mm}{
\begin{tabular}{cccccc}
\toprule[1pt]
Model & Image & LaRE & Method & AVG. ACC & AVG. AP \\ \hline
A     &     \checkmark        &            & None   & 73.1    & 92.3   \\
B     &             &     \checkmark       & None   & 66.2    & 68.3   \\
C     &       \checkmark      &    \checkmark        & Concat & 76.8    & 93.5   \\
D     &        \checkmark     &      \checkmark      & EGRE   & \textbf{84.5}    & \textbf{99.5} \\
\toprule[1pt]
\end{tabular}}
    \caption{Results of using different input. Models are trained on the SDv1.5 subset and tested on all the subsets. `AVG' indicates average scores over 8 subsets. `Concat' indicates concat LaRE with the image feature map after the last convolution block. }\smallskip
    \label{table3}
\end{table}

\subsection{Qualitative Results and Visualizations}
We utilize t-SNE\cite{van2008visualizing} visualization to illustrate feature vectors derived from the final layers of our model and the baseline model, as depicted in Fig.~\ref{figfs}. Each model is trained on the SDv1.5 subset of GenImage and evaluated across all eight generative methods. Although both models exhibit good feature discrimination on seen images, they both have limitations when it comes to unseen scenarios. For images from MJ, ADM, and BigGAN, both models show overlaps between real and fake images in the feature space. However, our model demonstrates a smaller proportion of overlap, indicating its better generalization ability.

\section{Conclusion}
In this paper, we propose a novel reconstruction-based diffusion-generated image detection method called La$\text{RE}^2$. We come up with LaRE, a novel and more efficient reconstruction-based feature by reconstructing the image in the Latent space. Notably, LaRE is 8 times faster compared with existing reconstruction-based methods. By incorporating LaRE with the Error-guided Feature Refinement module (EGRE). Our La$\text{RE}^2$ achieves superior performance on diffusion-generated image detection, demonstrating state-of-the-art performance. 

%%%%%%%%% REFERENCES
{\small
\bibliographystyle{ieee_fullname}
\bibliography{egbib}
}
\end{document}

\title{\LaTeX\ Guidelines for Author Response}  % **** Enter the paper title here

% \maketitle
\thispagestyle{empty}
\appendix

%R1
%The motivations behind the approach are validated well (e.g. Fig 1b and Fig 2) and address practical considerations such as speed
% Extensive ablations are provided which are valuable in understanding the trade-offs involved
% Strong results compared to baselines on the GenImage benchmark

%R2
% The idea of the work is simple and easy to follow.
% The organization of the paper is clear and the writing is well.
% Some techniques such as Spatial Attention and Channel Attention Strategies seems effective to improve performance.

% R3
% improve the ACC/AP in a great manner.
% The paper presents a comprehensive evaluation of the proposed framework, including quantitative and qualitative assessments. The authors provide a detailed description of the lare2 and the detailed ablation study. The evaluation results demonstrate the effectiveness of the proposed framework in diffusion-generated image detection.

%%%%%%%%% BODY TEXT - ENTER YOUR RESPONSE BELOW
% Our responses are listed below and all the additional experiment results will be added to our final version:
% Our responses to reviewers’ concerns are listed below:

% Q1: Presentation issues.

% A1: Thanks for comments. We will carefully check the presentation errors and correct them in our final version.

\setcounter{table}{3}
\begin{table}[t]
	 \centering
% \resizebox{0.48\textwidth}{!}{
\scriptsize
\renewcommand\arraystretch{1.0}
\setlength{\tabcolsep}{0.6mm}{

\begin{tabular}{lccccccccc}
\hline
\toprule[1pt]
\multirow{2}{*}{Model} & \multicolumn{8}{c}{Testing Diffusion Generators}       & \multirow{2}{*}{\begin{tabular}[c]{@{}l@{}}Total\\ Avg.\end{tabular}} \\ \cline{2-9}
                       & ADM  & DDPM & IDDPM & LDM  & PNDM & SDV1 & SDV2 & VQDM &                                                                       \\ \hline
CNNSplot {[}39{]}      & 100  & 99.6 & 100   & 99.1 & 85.8 & 98.2 & 99.9 & 96.1 & 97.3                                                                  \\
F3Net {[}29{]}         & 99.6 & 99.6 & 99.9  & 99.2 & 95.3 & 91.7 & 97.8 & 98.7 & 97.7                                                                  \\
DIRE {[}40{]}          & 100  & 100  & 100   & 100  & 100  & 100  & 100  & 100  & 100                                                                   \\
Ours                   & 100  & 99.4 & 100   & 99.9 & 99.9 & 100  & 99.9 & 100  & 99.9 \\\toprule[1pt]                                                           
\end{tabular}

}
% }
    \caption{Comparisons of AP on DiffusionForensics~[40] Dataset. }
    \label{table:DF}
    \vspace{-0.3cm}
\end{table}

\noindent \textbf{Necessity of diffusion-generated image detection}:
Diffusion-generated image detection, a subset of forgery detection, has unique features that necessitate special attention. The creation process leaves specific artifacts~\cite{corvi2023intriguing, corvi2023detection, wang2023dire} requiring specialized detection methods. The prevalence and realism of these images have increased due to the development of diffusion models~\cite{dhariwal2021diffusion, rombach2022high}, presenting new challenges~\cite{juefei2022countering}. Focusing on this task improves our understanding of various image forgeries.

% Indeed, diffusion-generated image detection can be seen as a subtask of the broader forgery detection task. However, it's important to note that it has unique characteristics that warrant specific attention. Firstly, the process of creating these images leaves distinct artifacts~[6, 7, 40] that require specialized detection methods. Secondly, with the rise of deep learning models, diffusion-generated images have become more prevalent and realistic~[9, 32], posing a new challenge in forgery detection~[16]. Lastly, by focusing on this task, we not only address this emerging issue but also contribute to the broader field by improving our understanding and detection of various types of image forgeries.

\noindent \textbf{More theoretical analysis on the difference between fakes and reals}:
Generated images have a higher posterior probability after being sampled from the generative model, making them more easily reconstructed by the model. The observed phenomenon is generalizable, as evidenced across eight GenImage\cite{zhu2023genimage} benchmark models.

\noindent\textbf{Experiments on DiffusionForensics}: In Tab.~\ref{table:DF}, we present the results on the DiffusionForensics\cite{wang2023dire} Dataset. Following DIRE\cite{wang2023dire}, we train our model on LSUN-B\cite{yu2015lsun}. The fake images are generated by ADM\cite{dhariwal2021diffusion}. Both our method and DIRE exhibit performance advantage over existing methods. Notably, our method not only matches DIRE in terms of effectiveness but also outperforms it in efficiency, which is \textbf{8x} faster. This clearly underscores the superior efficacy and efficiency of our approach.

% \scriptsize[1] Automated radiology report generation using conditioned transformer

%%%%%%%%% REFERENCES
{\small
\bibliographystyle{ieee_fullname}
\bibliography{egbib}
}